\documentclass{article}

\usepackage{microtype}
\usepackage{graphicx}
\usepackage{subcaption}
\usepackage{booktabs} % for professional tables

\usepackage{hyperref}

\usepackage[preprint]{icml2026}

\usepackage{amsmath}
\usepackage{amssymb}
\usepackage{mathtools}
\usepackage{amsthm}

\usepackage[capitalize,noabbrev]{cleveref}

\usepackage{url}
\usepackage{multirow}
\usepackage{enumitem}
\usepackage{stmaryrd}
\usepackage{array}
\usepackage{threeparttable}
\usepackage{blindtext}
\usepackage{amssymb}

\usepackage{svg}
\usepackage{placeins}

\usepackage{soul}
\usepackage[utf8]{inputenc}
\usepackage{graphicx} % 导入包
\usepackage{amsmath}
\usepackage{booktabs}
\usepackage{lipsum}

\usepackage{threeparttable}
\usepackage{makecell}
\usepackage{booktabs}
\usepackage{multirow}
\usepackage{marvosym}
\usepackage{tikz}
\usepackage{amsfonts,amssymb}
\usepackage{longtable}
\usepackage{float}
\usepackage{placeins}
\usepackage[table]{xcolor}

\newcommand{\hide}[1]{}

\newcommand{\dataset}{UltraVQA}
\usepackage{amsmath,amsfonts,bm}

\def\eqref#1{equation~\ref{#1}}
\def\1{\bm{1}}

\DeclareMathAlphabet{\mathsfit}{\encodingdefault}{\sfdefault}{m}{sl}
\SetMathAlphabet{\mathsfit}{bold}{\encodingdefault}{\sfdefault}{bx}{n}

\theoremstyle{plain}

\theoremstyle{definition}

\theoremstyle{remark}

\usepackage[textsize=tiny]{todonotes}

\begin{document}

\twocolumn[
  % \icmltitle{Submission and Formatting Instructions for \\
  %   International Conference on Machine Learning (ICML 2026)}
  \icmltitle{Analytic Score Optimization for Multi Dimension Video Quality Assessment}


  % \icmlsetsymbol{intern}{*}

  \begin{icmlauthorlist}
    \icmlauthor{Boda Lin$^*$}{ks}
    \icmlauthor{Yongjie Zhu$^\dagger$}{ks}
    \icmlauthor{Wenyu Qin}{ks}
    \icmlauthor{Meng Wang}{ks}
    \icmlauthor{Pengfei Wan}{ks}
  \end{icmlauthorlist}

  \icmlaffiliation{ks}{Kling Team, Kuaishou Technology, Beijing, China}
  \icmlcorrespondingauthor{Yongjie Zhu}{zhuyongjie@kuaishou.com}

  \vskip 0.3in
]

\newcommand\blfootnote[1]{%
  \begingroup
  \renewcommand\thefootnote{}\footnote{#1}%
  \addtocounter{footnote}{-1}%
  \endgroup
}

\blfootnote{$^*$This work was conducted during the author's internship at Kling Team, Kuaishou Technology. $^\dagger$Project learer.}

\printAffiliationsAndNotice{}  % no special notice (required even if empty)

\begin{abstract}
Video Quality Assessment~(VQA) is evolving beyond single-number mean opinion score toward richer, multi-faceted evaluations of video content.
In this paper, we present a large-scale multi-dimensional VQA dataset \dataset~ that encompasses diverse User-Generated Content~(UGC) annotated across five key quality dimensions: Motion Quality, Motion Amplitude, Aesthetic Quality, Content Quality, and Clarity Quality.
Each video in our dataset is scored by over 3 human raters on these dimensions, with fine-grained sub-attribute labels, and accompanied by an explanatory rationale generated by GPT based on the collective human judgments. 
To better leverage these rich annotations and improve discrete quality score assessment, we introduce Analytic Score Optimization (ASO), a theoretically grounded post-training objective derived for multi-dimensional VQA. 
By reframing quality assessment as a regularized decision-making process, we obtain a closed-form solution that naturally captures the ordinal nature of human ratings, ensuring alignment with human ranking preferences. 
In experiments, our method outperforms most baselines including closed-source APIs and open-source models, while also reducing mean absolute error (MAE) in quality prediction. 
Our work highlights the importance of multi-dimensional, interpretable annotations and reinforcement-based alignment in advancing video quality assessment.
\end{abstract}
    
\section{Introduction}
\label{sec:intro}

Assessing the perceptual quality of videos remains a long-standing challenge in computer vision and multimedia processing.
% Traditional video quality assessment (VQA) methods---including conventional no-reference metrics and early learning-based models---often compress quality into a single scalar, typically reported as the Mean Opinion Score (MOS).
Traditional VQA methods often reduce perceptual quality to a single scalar, most commonly the Mean Opinion Score (MOS), whether they are hand-crafted no-reference metrics or early learning-based models.
While convenient for benchmarking and optimization, a single score is inherently limited: it obscures why a video is perceived as good or bad, and cannot disentangle the diverse factors that jointly shape viewers' quality of experience~\cite{mvqa-68k}.
As user-generated content (UGC) becomes increasingly diverse in style, capture conditions, and post-processing, there is a growing demand for multi-dimensional VQA that evaluates quality along multiple, more interpretable axes~\cite{duan2025finevq,videoscore2}.

Recent work has begun to move in this direction by decomposing video quality into multiple aspects such as spatial quality, temporal consistency, and aesthetics~\cite{md-vqa,duan2025finevq,wu2023dover,videoscore2,mvqa-68k}.
For example, MD-VQA~\cite{md-vqa} studies semantic, distortion, and motion factors for UGC livestream videos, while MVQA~\cite{mvqa-68k} emphasizes multi-factor quality understanding with richer annotations.
More recently, vision-language models (VLMs) have been applied to VQA, bringing stronger high-level reasoning and the ability to produce textual explanations beyond pixel-level cues~\cite{mvqa-68k,videoscore2}.
% \lin{modify}
Despite this progress, existing VLM-based VQA systems still face two practical limitations.
First, models often yield coarse overall judgments and are insufficiently sensitive to nuanced quality factors (e.g., subtle motion artifacts or aesthetic trade-offs), especially under distribution shifts.
Second, in the absence of carefully curated training data with consistent multi-dimensional labels and rationale supervision, even strong VLMs can exhibit weak correlation with human ratings on standard benchmarks (citation to be added).

To address these challenges, we introduce a large-scale VQA dataset, \dataset, designed for both comprehensiveness and interpretability.
% \dataset~ characterizes video quality using five core dimensions that cover low-level and high-level aspects:
\dataset~ characterizes video quality using five core dimensions:
(1) \textbf{Motion Quality}, capturing temporal smoothness and stability;
(2) \textbf{Motion Amplitude}, measuring the degree and extent of motion;
(3) \textbf{Aesthetic Quality}, reflecting perceived visual appeal such as composition and lighting;
(4) \textbf{Content Quality}, assessing semantic relevance and meaningfulness; and
(5) \textbf{Clarity Quality}, covering sharpness, resolution, noise, and compression artifacts.
Each dimension further includes fine-grained sub-attributes, enabling models to learn structured quality judgments rather than conflating heterogeneous factors into a single score.

Beyond scores and sub-attributes, \dataset~ also provides explanatory rationales for improved transparency.
We employ 40 specially trained annotators, and each video clip is scored by at least three annotators across all five dimensions, together with detailed attribution tags.
We further use GPT-4.1 to synthesize concise rationale paragraphs that explain the assigned scores based on the human-provided ratings and tags.
% (optionally followed by human verification; details in Sec.~X).
This rationale supervision is intended to encourage models to not only predict scores but also justify their evaluations.

In multi-dimensional VQA, the supervision signal is fundamentally ordinal and discrete: each dimension is rated on a small set of levels (e.g., 1.0–5.0 with 0.5 steps used in our \dataset), and the evaluation emphasizes rank or linear agreement rather than exact regression to a continuous MOS. 
However, most post-training pipelines treat scoring either as (i) free-form generation with sparse correctness rewards, or (ii) continuous regression, both of which ignore the finite, ordered label space and often yield poorly calibrated score distributions. 
This mismatch becomes particularly problematic for motion-centric dimensions, where the perceived difference between adjacent levels can be subtle and policy-gradient updates are high-variance. 
Motivated by this, we introduce Analytic Score Optimization (ASO): instead of relying on stochastic policy gradients, we formulate discrete scoring as a KL-regularized one-step bandit and derive a closed-form optimal score policy over discrete levels. 
ASO turns discrete score alignment into a stable, sample-efficient soft-target learning objective, which complements standard RL alignment and directly exploits the structure of ordinal score spaces.

% On top of \dataset, we develop \textbf{Analytic Score Optimization (ASO)} for discrete quality scoring.
% Unlike generic reinforcement learning post-training that relies on stochastic policy gradients, ASO leverages the finite label space of discrete quality scores.
% From a reinforcement learning perspective, we formulate discrete scoring as an optimization problem and derive an analytic objective, resulting in a stable and sample-efficient training procedure.

We validate our approach on \dataset~ as well as multiple public benchmarks.
Experiments show that VLM trained with ASO achieves improvements over strong baselines, and that training with rationale supervision yields more faithful explanations and improved performance.
% Notably, fine-tuning on \dataset~ enables a 7B VLM to outperform larger models on our test set under standard evaluation protocol.

In summary, our contributions are:
\begin{itemize}
    \item \textbf{Dataset:} We construct \dataset, a large-scale multi-dimensional VQA dataset with five quality dimensions and fine-grained sub-attributes. Each clip is annotated by multiple trained annotators and accompanied by rationale explanations synthesized from human annotations.
    \item \textbf{Method:} We introduce Analytic Score Optimization (ASO), an RL-inspired analytic optimization objective tailored to discrete quality scoring with a finite label space.
    \item \textbf{Results:} We demonstrate consistent gains on \dataset~ and public benchmarks, and show that rationale supervision improves interpretability and zero-shot generalization.
\end{itemize}

\section{Related Work}
\label{sec:related}

\subsection{VQA Datasets}
Over the past decade, numerous VQA datasets have been established to advance video quality assessment. 
Early datasets~\cite{ghadiyaram2017capture,nuutinen2016cvd2014,sinno2018large,wang2016mcl,wang2017videoset,wang2021rich,yu2021predicting} often used limited source videos with synthetic distortions like compression and transmission errors.
Later “in-the-wild” UGC datasets such as KoNViD-1k~\cite{hosu2017konstanz}, YouTube-UGC~\cite{wang2019youtube}, and LSVQ~\cite{ying2021patch} collected real-world videos with diverse authentic distortions.
These benchmarks typically provide a single MOS per video as the ground-truth quality. 
While valuable, single-score datasets have inherent limitations that they do not reveal which aspect of quality influenced the rating, and models trained on them can struggle to generalize across content types~\cite{duan2025finevq, md-vqa, liu2022quality, chen2021unsupervised, wu2023towards}.
Recognizing the multi-faceted nature of video quality, researchers have begun exploring multi-dimensional VQA. 
MD-VQA~\cite{md-vqa} introduced a concept of multi-dimensional quality for UGC live videos, focusing on three dimensions (semantic quality, distortion, motion) and even proposed an interpretable model that fuses features corresponding to these factors.
Their results showed better performance than one-dimensional approaches, confirming the benefit of multi-dim labels.
Our work extends this line by covering a broader set of quality dimensions, including subjective aesthetics and content coherence, and by scaling up to a much larger dataset. 
Most closely related to our dataset is the very recent MVQA-68K benchmark~\cite{mvqa-68k}, which independently also emphasizes multi-dimensional annotation (seven dimensions including aesthetics, motion, etc.) and provides chain-of-thought reasoning for interpretability. 
Our dataset differs in the specific taxonomy of quality dimensions (we define five major categories tailored to UGC, whereas MVQA-68K uses seven slightly different ones) and in the scale of human annotation per video (30+ ratings each, enabling robust consensus).

\subsection{VQA Methods}
Traditional VQA methods can be classified into full-reference (FR), reduced-reference, and no-reference (NR) methods. 
FR methods (e.g., PSNR, SSIM, VMAF) assume access to an undistorted reference video and compute differences, which is impractical for UGC. 
NR methods rely on learning-based features or hand-crafted features to predict quality directly from the distorted video. Before the deep learning era, NR-VQA models used features targeting distortions like blockiness or blur. 
More recent deep learning models (e.g., VSFA, TLVQM, PVQ, etc.) learn to regress MOS from deep CNN or CNN with LSTM features of video frames. However, these approaches usually still output a single score per video, lacking interpretability. 
The NR-VQA model in MD-VQA~\cite{md-vqa}, for instance, improved performance by explicitly modeling multiple factors (using separate CNNs for semantic, distortion, motion features and fusing them), but its output remained an overall MOS, and the interpretability came from the model architecture rather than human-readable explanations.

\subsection{VLM and Post-training}
Due to their strong generalization and instruction-following capability, Vision-Language Models (VLMs) are increasingly adopted as versatile backbones for VQA task.
Representative frameworks (e.g., LLaVA~\cite{liu2023llava}, BLIP-2~\cite{li2023blip2}, Flamingo~\cite{alayrac2022flamingo}, Kosmos-2~\cite{peng2023kosmos2}) demonstrate effective recipes for coupling vision encoders with LLMs to enable multimodal reasoning and structured generation.
Recent open-source families such as Qwen-VL~\cite{qwen3vl,qwen25vl,qwen2vl}, MiniCPM-V~\cite{minicpmv45}, and InternVL~\cite{intern3vl,intern25vl} further strengthen accessibility and practical deployment.

Compared to conventional NR-VQA regressors, VLM-based scorers can incorporate higher-level semantic cues (e.g., content coherence) and provide more interpretable outputs. Nevertheless, reliable quality scoring requires post-training: models must learn (i) strict output formats, (ii) calibrated discrete scoring behaviors, and (iii) alignment with human rating criteria across diverse content domains.

Direct Preference Optimization (DPO)~\cite{rafailov2023dpo} is a widely adopted method.
Subsequent works propose reference-free or simplified variants, such as ORPO~\cite{hong2024orpo} and SimPO~\cite{meng2024simpo}, as well as objectives that learn from weaker binary signals, e.g., KTO~\cite{ethayarajh2024kto}.
More recently, Implicit Preference Optimization (IPO) views LMs as implicit preference classifiers, further reducing reliance on explicit reward modeling~\cite{garg2025ipo}.
These approaches are particularly attractive for VQA settings where supervision can be naturally expressed as pairwise preferences (e.g., which video looks better) or ordinal levels.

RLHF typically optimizes a learned reward model with online policy updates, where PPO is a standard choice due to its stability~\cite{schulman2017proximal}.
However, online RL can be memory- and compute-intensive, motivating more efficient alternatives such as GRPO~\cite{shao2024deepseekmath} in large-scale practice.

\section{Dataset}
\label{sec:dataset}

\begin{figure*}[t]
\centering
\includegraphics[width=1.0\linewidth]{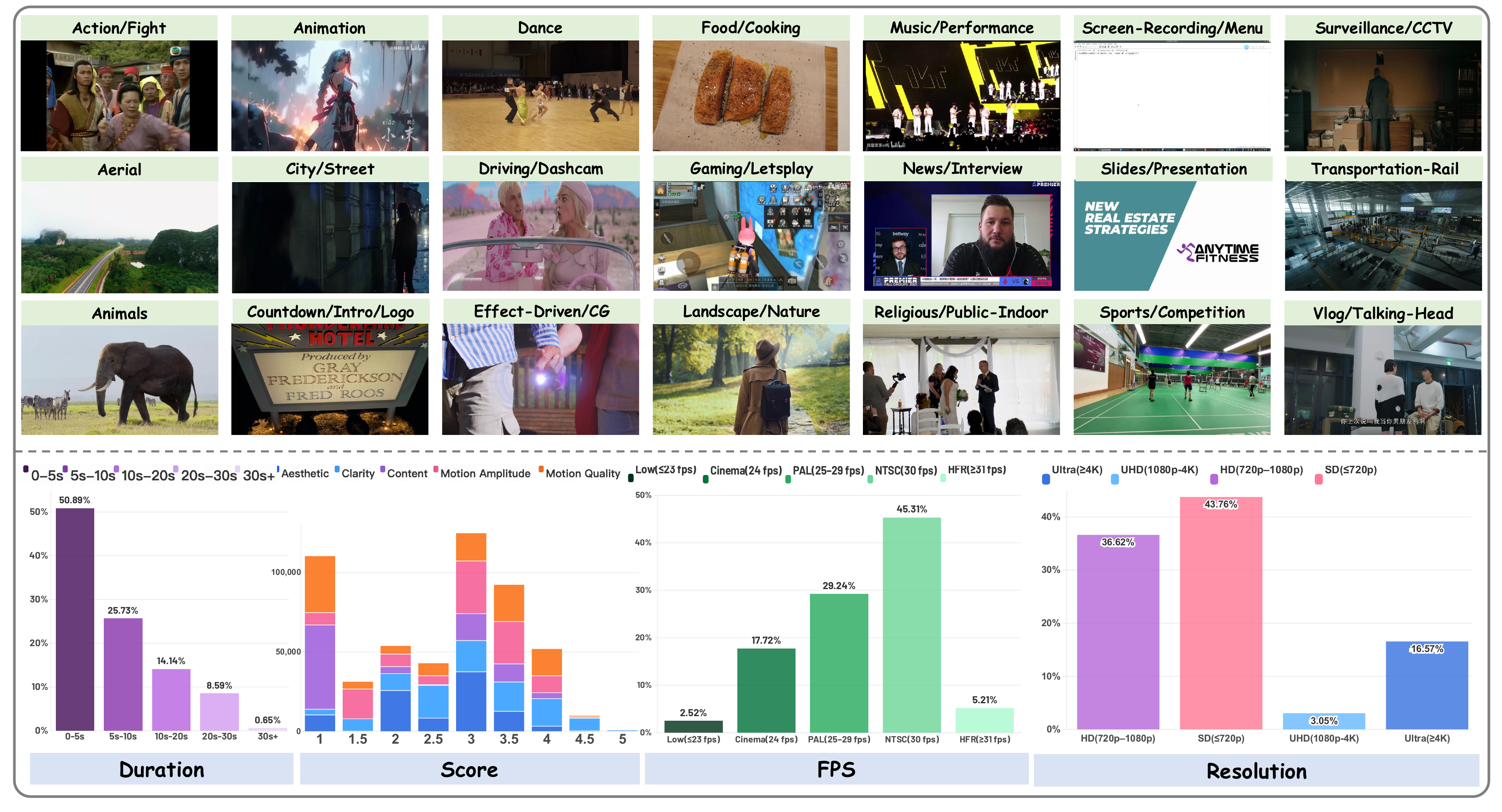}
\caption{
Cases for each category and distribution of statistics of our \dataset.
}
\label{fig:data:statis}
\end{figure*}
\subsection{Data Collection}
\dataset~ contains approximately 40,000 video clips curated from a mixture of in-the-wild UGC platforms and professional content sources, covering diverse genres such as vlogs, sports, gaming, education, interviews, and cinematic edits. 

We ensured diversity in the footage: videos include both camera-captured authentic content and some AI-generated or edited content, to reflect modern video sources. 
We also intentionally included a subset of professionally produced or artistically edited high-aesthetic clips to counterbalance the fact that many widely used VQA datasets are dominated by imperfect/distorted UGC and contain relatively few truly high-quality samples~\cite{hosu2017konstanz,ying2021patch,gotz2021konvid,sinno2018large}.

\subsection{Quality Dimensions and Sub-attributes}
We annotate five complementary quality dimensions: Motion Quality (temporal smoothness and stability), Motion Amplitude (degree/extent of motion), Aesthetic Quality (composition, color, lighting, and overall visual appeal), Content Quality (semantic coherence, informativeness, and subject completeness), and Clarity Quality (sharpness, resolution, noise, and compression artifacts). 
For each dimension, raters additionally select fine-grained sub-attribute tags from a curated checklist (e.g., camera shake, motion blur, over-compression, poor lighting). 
These tags enable more structured error analysis and serve as grounded evidence for rationale synthesis.
The details of tags are shown in Appendix~\ref{app:sec:data}.

\subsection{Data Statistics}
The statistical details of our \dataset~ are shown in the Figure~\ref{fig:data:statis}.
Considering the current processing power of VLM and the limitations of GPU memory, we divide most of the videos into segments of less than 30 seconds, and retained a small number of videos longer than 30 seconds.
To ensure the \dataset~ covers as many visual quality levels as possible, we select videos ranging from 480p to 4K resolution. 
Notably, videos at 4K and higher resolutions comprise $16.5\%$ of our \dataset.
We maintain the original FPS for all videos to the greatest extent possible, covering a variety of common video standards, including Cinema, PAL, NTSC, etc.
In addition, the average number of annotation tokens is 94.75 and the average number of annotation words for each video is 76.95.
Finally, we analyzed the thematic range of the videos, as shown in the Figure~\ref{fig:data:statis}, which includes 16 major categories.

\subsection{Data Annotation}
\begin{figure*}[t]
\centering
\includegraphics[width=1.0\linewidth]{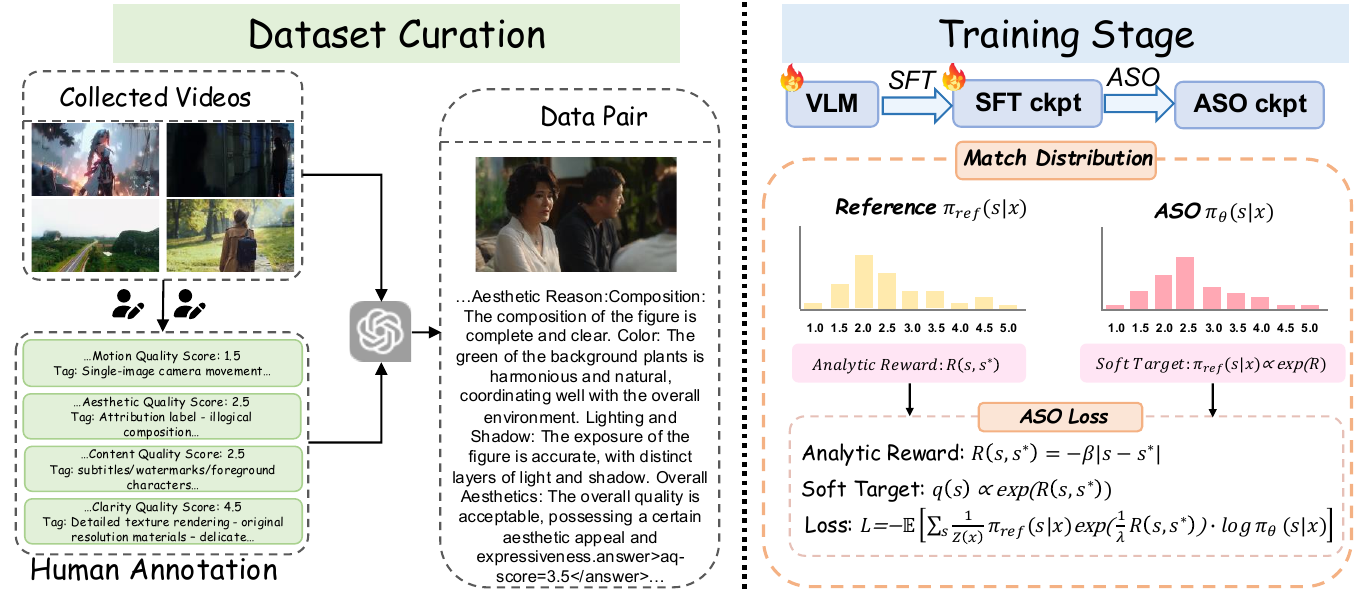}
\caption{
Overview of our method. We curate data and human annotators to annotate video quality scores and quality tags with GPT generated reasons. Then we use 2-stage training to elicit the model’s thinking ability before scoring.
}
\label{fig:model}
\end{figure*}

The dataset annotation process is shown in the Figure~\ref{fig:model}. 

To capture the multifaceted nature of User-Generated Content (UGC), we established a hierarchical annotation taxonomy covering both high-level quality dimensions and fine-grained attribution tags.

Quality Dimensions and Fine-grained Attributes. We define five complementary quality dimensions: Motion Quality, Motion Amplitude, Aesthetic Quality, Content Quality, and Clarity Quality. Unlike previous datasets that rely solely on holistic scoring, we require annotators to provide grounded evidence for their ratings. For each dimension, annotators must select applicable sub-attributes from a curated checklist of failure modes and highlights. For instance, Clarity Quality includes tags such as Blur, Noise/Grain, and Compression Artifacts, while Aesthetic Quality covers attributes like Composition, Lighting, and Color Grading. These categorical tags serve a dual purpose: they enable structured error analysis and act as ``anchor points" for generating hallucinogen-free textual rationales.

Robust Annotation Protocol. We engaged a pool of 40 trained professional annotators to ensure consistency. The annotation process follows a rigorous redundancy protocol where each video clip is independently evaluated by at least three annotators. Raters assess each dimension on a 5-point Likert scale (1.0 to 5.0, with 0.5 intervals), accompanied by the selection of relevant tags. To mitigate subjective bias inherent in UGC assessment, we implement a post-processing filtering step: annotations with excessive variance or inconsistency are flagged and re-evaluated. The final ground-truth score for each dimension is derived through statistical aggregation (typically the mean) of valid ratings, ensuring a robust distribution aligned with human perception.

Evidence-Grounded Rationale Synthesis. Beyond numerical scores, UltraVQA provides interpretable textual supervision. Directly collecting free-form comments from crowd-workers often yields noisy or sparse text. Instead, we propose an evidence-grounded synthesis pipeline. We aggregate the human-selected sub-attribute tags and raw comments to form a structured ``evidence set." We then prompt GPT-4.1 to synthesize a concise, coherent rationale conditioned strictly on: (i) the visual content, (ii) the aggregated ground-truth score, and (iii) the collective human evidence. This approach distills the consensus of multiple human raters into high-quality natural language, ensuring that the generated rationales are not model hallucinations but are faithfully grounded in human judgment.

\section{Method}
\label{sec:method}

Consistent with common post-training practice, we first perform supervised fine-tuning (SFT) on a vision-language model (VLM) to equip it with reliable formatted outputs and a basic capability for video quality scoring. We then apply GRPO on top of the SFT checkpoint to further improve performance. 
Standard regression losses (e.g., MSE) treat human ratings as deterministic point estimates, ignoring the inherent subjectivity and ambiguity in visual quality perception. We argue that an ideal VQA model should learn an optimal score distribution rather than a single value. ASO achieves this by balancing alignment with the ground truth against a KL-divergence constraint, effectively modeling the uncertainty in human judgment.

\subsection{SFT and Post-training Pipeline}
\label{sec:sft_pipeline}

We adopt a standard two-stage post-training pipeline to align the Vision-Language Model (VLM) for the VQA task. 
\textbf{Stage I (Supervised Fine-Tuning).} We first perform SFT to equip the VLM with the capability to follow instructions and generate valid formatted outputs. The model is trained to output five dimension-wise scores along with a short rationale, optimized via a standard cross-entropy loss over the target response.

\textbf{Stage II (Preference Alignment via GRPO).} 
Starting from the SFT checkpoint, we further train the model with popular post-training. 
To establish a strong baseline consistent with state-of-the-art RLHF practices, we explicitly employ \textbf{Group Relative Policy Optimization (GRPO)}~\cite{shao2024deepseekmath}. 
Specifically, for each input video and dimension, we sample a group of $G$ candidate outputs $\{o_1, ..., o_G\}$ from the old policy and optimize the model based on their relative rewards. This stage allows us to validate the effectiveness of standard reinforcement learning before introducing our analytic approach.

\textbf{Reward Formulation.} 
Crucial to the alignment stage (both for the GRPO baseline and our proposed ASO) is the design of the reward function. To capture both the discrete accuracy and the continuous ordinal nature of the scores, we define a composite scalar reward consisting of an \textit{Accuracy Reward} and a \textit{Distribution Reward}:

\begin{itemize}
    \item \textbf{Accuracy Reward ($R_{acc}$):} A discrete reward that encourages the model to hit the exact ground-truth bin.
    \begin{equation}
        R_{acc} =
        \begin{cases}
            1.0 & \text{if } \text{round}(pred) = \text{round}(gt) \\
            0.0 & \text{otherwise}
        \end{cases}
    \end{equation}
    
    \item \textbf{Distribution Reward ($R_{dist}$):} A continuous reward that penalizes the absolute deviation, guiding the model towards the ground truth even when the exact match is missed.
    \begin{equation}
        R_{dist} = 5.0 - \left\vert pred - gt \right\vert
    \end{equation}
\end{itemize}

The final reward is a weighted sum of these components, optionally combined with a format reward to ensure structural validity.

\subsection{Analytic Score Optimization}
\textbf{Problem formulation.}
We consider a single quality dimension (e.g., motion quality) where each video
\(x\) is annotated with a scalar target score \(s^*(x)\) in a discrete set
\(\mathcal{S} = \{s_1, \dots, s_{|\mathcal{S}|}\}\) (e.g., \(\{1.0,1.5,\dots,5.0\}\)).
A policy \(\pi_\theta(s \mid x)\) parameterized by \(\theta\) assigns a
probability distribution over candidate scores \(s \in \mathcal{S}\).
Taking an action corresponds to choosing a score from \(\mathcal{S}\).

We define a reward function \(R(s, s^*(x))\) which evaluates how suitable
a predicted score \(s\) is for the ground-truth score \(s^*(x)\).
Typical choices include distance-based rewards such as
\(R(s, s^*) = -\lvert s - s^* \rvert\) or \(R(s, s^*) = -(s-s^*)^2\).

\textbf{Regularized bandit objective.}
For this discrete scoring problem, the interaction with each video \(x\)
is a one-step bandit: the policy chooses a single score and immediately
receives a scalar reward. A natural RL objective in this setting is to
maximize the expected reward under a KL-regularized policy:
\begin{equation}
\begin{aligned}
  F(\pi) &=
  \mathbb{E}_{x \sim \mathcal{D}}
  [
    \sum_{s \in \mathcal{S}} \pi(s \mid x)\, R(s, s^*(x)) \\
    &\quad 
    - \lambda\, \mathrm{KL}\!\left(
      \pi(\cdot \mid x)\, \big\| \,
      \pi_{\text{ref}}(\cdot \mid x)
    \right)
  ],
  \label{eq:aso-objective}
\end{aligned}
\end{equation}
where \(\mathcal{D}\) is the data distribution over videos,
\(\pi_{\text{ref}}(s \mid x)\) is a fixed reference policy
(e.g., an SFT model or a simple prior), and \(\lambda > 0\) controls how
strongly the policy is regularized towards \(\pi_{\text{ref}}\).
The KL term prevents the policy from drifting too far from the reference
distribution when the reward signal is sparse or noisy, which is standard
in RLHF-style regularized policy optimization.

Since the expectation over \(x\) decomposes, we can maximize
\(F(\pi)\) by independently optimizing, for each \(x\), the inner functional
\begin{equation}
\begin{aligned}
  F_x(\pi) &=
  \sum_{s \in \mathcal{S}} \pi(s \mid x)\, R(s, s^*(x)) \\
  &- \lambda\, \mathrm{KL}\!\left(
    \pi(\cdot \mid x)\, \big\| \,
    \pi_{\text{ref}}(\cdot \mid x)
  \right),
  \label{eq:aso-objective-single-x}
\end{aligned}
\end{equation}
over all distributions \(\pi(\cdot \mid x)\) on the simplex.

\textbf{Closed-form optimal score policy.}
For a fixed video \(x\), Eq.~\eqref{eq:aso-objective-single-x} is a convex
optimization problem over the discrete simplex. Introducing a Lagrange
multiplier \(\alpha(x)\) to enforce the normalization
\(\sum_{s \in \mathcal{S}} \pi(s \mid x) = 1\), we write the Lagrangian
\begin{equation}
\begin{aligned}
  \mathcal{L}_x(\pi, \alpha)
  &=
  \sum_{s \in \mathcal{S}} \pi(s \mid x)\, R(s, s^*(x))
  \\
  &\quad
  - \lambda
  \sum_{s \in \mathcal{S}}
  \pi(s \mid x)
  \log \frac{\pi(s \mid x)}{\pi_{\text{ref}}(s \mid x)}
  \\
  &\quad
  + \alpha(x)
  \left(
    \sum_{s \in \mathcal{S}} \pi(s \mid x) - 1
  \right).
\end{aligned}
\label{eq:aso-lagrangian}
\end{equation}

Taking the derivative of \(\mathcal{L}_x\) w.r.t.\ \(\pi(s \mid x)\) and
setting it to zero yields, for each \(s \in \mathcal{S}\),
\begin{equation}
\begin{aligned}
  0
  &=
  \frac{\partial \mathcal{L}_x}{\partial \pi(s \mid x)}
  \\
  &=
  R(s, s^*(x))
  - \lambda
    \left(
      \log \frac{\pi(s \mid x)}{\pi_{\text{ref}}(s \mid x)} + 1
    \right)
  + \alpha(x).
\end{aligned}
\label{eq:aso-stationary}
\end{equation}
Rearranging Eq.~\eqref{eq:aso-stationary} gives
\begin{equation}
  \log \pi(s \mid x)
  =
  \log \pi_{\text{ref}}(s \mid x)
  + \frac{1}{\lambda} R(s, s^*(x))
  + C(x),
  \label{eq:aso-log-policy}
\end{equation}
where \(C(x)\) collects terms independent of \(s\).
Exponentiating and normalizing over \(\mathcal{S}\) leads to the
closed-form optimal policy
\begin{equation}
  \pi^*(s \mid x)
  =
  \frac{1}{Z(x)}\,
  \pi_{\text{ref}}(s \mid x)\,
  \exp\!\left(\frac{1}{\lambda} R(s, s^*(x))\right),
  s \in \mathcal{S},
  \label{eq:aso-optimal-policy}
\end{equation}
with the partition function
\begin{equation}
  Z(x)
  =
  \sum_{s' \in \mathcal{S}}
  \pi_{\text{ref}}(s' \mid x)\,
  \exp\!\left(\frac{1}{\lambda} R(s', s^*(x))\right).
  \label{eq:aso-normalizer}
\end{equation}
Eq.~\eqref{eq:aso-optimal-policy} shows that the optimal score policy
reweights the reference policy by a Boltzmann factor of the reward.

\textbf{Parametric approximation.}
In practice, we use a neural network to represent a parametric policy
\(\pi_\theta(s \mid x)\). Instead of directly maximizing
Eq.~\eqref{eq:aso-objective} by sampling, we treat the closed-form
\(\pi^*(s \mid x)\) in Eq.~\eqref{eq:aso-optimal-policy} as an
\emph{ideal teacher} distribution and train \(\pi_\theta\) to imitate it.
Concretely, we minimize the KL divergence from \(\pi^*\) to \(\pi_\theta\):
\begin{equation}
  L_{\text{score}}(\theta)
  =
  \mathbb{E}_{x \sim \mathcal{D}}
  \Big[
    \mathrm{KL}\!\big(
      \pi^*(\cdot \mid x)
      \,\big\|\,
      \pi_\theta(\cdot \mid x)
    \big)
  \Big].
  \label{eq:aso-kl-loss}
\end{equation}
Expanding the KL divergence yields
\begin{equation}
\begin{aligned}
  L_{\text{score}}(\theta)
  &=
  \mathbb{E}_{x \sim \mathcal{D}}
  \left[
    \sum_{s \in \mathcal{S}}
    \pi^*(s \mid x)
    \log
    \frac{
      \pi^*(s \mid x)
    }{
      \pi_\theta(s \mid x)
    }
  \right]
  \\
  &=
  \mathbb{E}_{x \sim \mathcal{D}}
  \left[
    \sum_{s \in \mathcal{S}}
    \pi^*(s \mid x)
    \log \pi^*(s \mid x)
  \right]
  \\
  &\quad
  -
  \mathbb{E}_{x \sim \mathcal{D}}
  \left[
    \sum_{s \in \mathcal{S}}
    \pi^*(s \mid x)
    \log \pi_\theta(s \mid x)
  \right].
\end{aligned}
\label{eq:aso-kl-expanded}
\end{equation}
The first term in Eq.~\eqref{eq:aso-kl-expanded} does not depend on
\(\theta\) and can be treated as a constant. Dropping this constant yields
the practical training loss
\begin{equation}
  \tilde{L}_{\text{score}}(\theta)
  =
  -\mathbb{E}_{x \sim \mathcal{D}}
  \left[
    \sum_{s \in \mathcal{S}}
    \pi^*(s \mid x)
    \log \pi_\theta(s \mid x)
  \right],
  \label{eq:aso-ce-loss}
\end{equation}
Substituting into Equation 10, we can obtain the final loss that depends on both the Reward and the reference model.
\begin{equation}
\label{eq:aso-ce-loss}
\begin{aligned}
\tilde{L}_{\text{score}}(\theta)
&=
-\mathbb{E}_{x \sim \mathcal{D}}
\Bigg[
\sum_{s \in \mathcal{S}}
\frac{1}{Z(x)}\,
\pi_{\text{ref}}(s \mid x)\,
\\
&\exp\!\left(\frac{1}{\lambda} R(s, s^*(x))\right)
\cdot
\log \pi_\theta(s \mid x)
\Bigg].
\end{aligned}
\end{equation}

which is exactly a cross-entropy loss with \emph{soft targets}
\(\pi^*(s \mid x)\) shaped by the reward and the reference policy.
Although Eq.~\eqref{eq:aso-ce-loss} looks identical to a standard SFT loss,
the target distribution \(\pi^*\) is the closed-form solution to the
regularized RL objective in Eq.~\eqref{eq:aso-objective}, rather than an
ad hoc label smoothing heuristic.

Crucially, while the reward $R$ is calculated based on the score tokens, the KL-divergence term $D_{KL}(\pi || \pi_{\text{ref}})$ acts as a regularizer for the entire sequence. This ensures that the generated rationales remain semantically coherent and faithful to the pre-trained knowledge of the SFT model, preventing the 'reward hacking' phenomenon where models generate gibberish to maximize scores.
\section{Experiment}
\label{sec:exp}

\begin{table*}[t]
\centering
\scalebox{0.6}{
\begin{tabular}{l
*{5}{cccc}
}
\toprule
\multirow{2}{*}{\textbf{Model}} &
\multicolumn{4}{c}{\textbf{Motion Quality}} &
\multicolumn{4}{c}{\textbf{Motion Amplitude}} &
\multicolumn{4}{c}{\textbf{Aesthetic Quality}} &
\multicolumn{4}{c}{\textbf{Content Quality}} &
\multicolumn{4}{c}{\textbf{Clarity Quality}} \\
\cmidrule(lr){2-5}\cmidrule(lr){6-9}\cmidrule(lr){10-13}\cmidrule(lr){14-17}\cmidrule(lr){18-21}
& \textbf{Acc} & \textbf{MAE} & \textbf{SRCC} & \textbf{PLCC}
& \textbf{Acc} & \textbf{MAE} & \textbf{SRCC} & \textbf{PLCC}
& \textbf{Acc} & \textbf{MAE} & \textbf{SRCC} & \textbf{PLCC}
& \textbf{Acc} & \textbf{MAE} & \textbf{SRCC} & \textbf{PLCC}
& \textbf{Acc} & \textbf{MAE} & \textbf{SRCC} & \textbf{PLCC} \\
\midrule
\rowcolor{gray!10}
\multicolumn{1}{l}{\textit{VLM-APIs}} & & & & & & & & & & & & & & & & & & & \\
\makecell[l]{GPT-4.1}
& $48.0\%$ & $1.151$ & $0.418$ & $0.386$
& $66.8\%$ & $0.595$ & $0.697$ & $0.693$
& $56.9\%$ & $0.744$ & $0.611$ & $0.615$
& $52.5\%$ & $0.992$ & $0.093$ & $0.096$
& $75.3\%$ & $0.514$ & $0.848$ & $0.839$ \\
\makecell[l]{Gemini-2.5Pro}
& $57.0\%$ & $0.849$ & $0.608$ & $0.590$
& $49.7\%$ & $0.888$ & $0.744$ & $0.723$
& $56.0\%$ & $0.812$ & $0.654$ & $0.643$ 
& $40.3\%$ & $1.138$ & $0.322$ & $0.277$
& $65.8\%$ & $0.611$ & $0.850$ & $0.824$ \\
\midrule
\rowcolor{gray!10}
\multicolumn{1}{l}{\textit{Open-source VLMs}} & & & & & & & & & & & & & & & & & & & \\
\makecell[l]{Qwen2.5-VL}
& $38.0\%$ & $1.233$ & $0.532$ & $0.536$
& $44.2\%$ & $1.042$ & $0.749$ & $0.733$
& $41.0\%$ & $1.187$ & $0.373$ & $0.361$
& $43.3\%$ & $1.220$ & $0.093$ & $0.097$
& $54.4\%$ & $0.786$ & $0.793$ & $0.799$ \\
\makecell[l]{MiniCPMV-4.5}
& $46.8\%$ & $1.157$ & $0.413$ & $0.425$
& $62.2\%$ & $0.704$ & $0.651$ & $0.648$
& $44.3\%$ & $1.052$ & $0.389$ & $0.394$
& $49.7\%$ & $1.018$ & $0.077$ & $0.099$
& $47.1\%$ & $0.900$ & $0.775$ & $0.774$ \\
\makecell[l]{VideoLLaMA3}
& $40.2\%$ & $1.353$ & $0.301$ & $0.362$
& $38.8\%$ & $1.140$ & $0.641$ & $0.615$
& $50.2\%$ & $0.973$ & $0.452$ & $0.490$
& $51.8\%$ & $1.006$ & $0.205$ & $0.157$
& $41.4\%$ & $1.010$ & $0.755$ & $0.771$ \\
\makecell[l]{VideoChat-Flash}
& $36.4\%$ & $1.468$ & $0.124$ & $0.144$
& $44.1\%$ & $1.132$ & $0.154$ & $0.117$
& $43.5\%$ & $1.121$ & $0.206$ & $0.124$
& $29.6\%$ & $1.380$ & $0.188$ & $0.183$
& $36.4\%$ & $1.082$ & $0.643$ & $0.574$ \\
\makecell[l]{InternVL-3.5}
& $41.0\%$ & $1.466$ & $0.205$ & $0.243$
& $71.7\%$ & $0.564$ & $0.771$ & $0.743$
& $40.3\%$ & $1.192$ & $0.508$ & $0.505$
& $40.1\%$ & $1.322$ & $0.141$ & $0.158$
& $40.9\%$ & $0.974$ & $0.817$ & $0.860$ \\
\makecell[l]{InternVideo-2.5}
& $35.5\%$ & $1.521$ & $0.091$ & $0.068$
& $49.5\%$ & $0.848$ & $0.430$ & $0.428$
& $45.7\%$ & $1.026$ & $0.308$ & $0.296$
& $46.1\%$ & $1.041$ & $0.028$ & $0.051$
& $50.7\%$ & $0.835$ & $0.685$ & $0.685$ \\
\midrule
\rowcolor{gray!10}
\multicolumn{1}{l}{\textit{VQA Baselines}} & & & & & & & & & & & & & & & & & & & \\
\makecell[l]{FineVQ}
& $61.3\%$ & $0.789$ & $0.691$ & $0.685$
& $69.2\%$ & $0.824$ & $0.705$ & $0.792$
& $72.5\%$ & $0.885$ & $0.743$ & $0.731$
& $58.4\%$ & $0.832$ & $0.682$ & $0.621$
& $74.1\%$ & $0.998$ & $0.782$ & $0.767$ \\
\makecell[l]{Q-Align}
& $34.4\%$ & $1.466$ & $0.260$ & $0.261$
& $43.2\%$ & $1.073$ & $0.199$ & $0.183$
& $44.3\%$ & $0.975$ & $0.616$ & $0.602$
& $24.7\%$ & $1.613$ & $0.235$ & $0.247$
& $36.5\%$ & $0.942$ & $0.876$ & $0.871$ \\
\makecell[l]{VideoScoreV2}
& $69.8\%$ & $0.767$ & $0.703$ & $0.725$
& $70.5\%$ & $0.505$ & $0.812$ & $0.801$
& $63.7\%$ & $1.072$ & $0.651$ & $0.639$
& $60.2\%$ & $0.818$ & $0.689$ & $0.628$
& $75.3\%$ & $0.881$ & $0.791$ & $0.776$ \\
\midrule
\rowcolor{gray!10}
\multicolumn{1}{l}{\textit{Ours}} & & & & & & & & & & & & & & & & & & & \\
\makecell[l]{SFT}
& $71.7\%$ & $0.622$ & $0.694$ & $0.710$
& $80.3\%$ & $0.419$ & $0.851$ & $0.847$
& $74.0\%$ & $0.467$ & $0.752$ & $0.766$
& $56.7\%$ & $0.915$ & $0.603$ & $0.562$
& $83.4\%$ & $0.428$ & $0.867$ & $0.859$ \\
\makecell[l]{GRPO}
& $81.0\%$ & $0.446$ & $0.806$ & $0.801$
& $90.4\%$ & $0.296$ & $0.921$ & $0.907$
& $84.7\%$ & $0.352$ & $\textbf{0.836}$ & $\textbf{0.841}$
& $69.0\%$ & $0.572$ & $0.790$ & $0.734$
& $85.0\%$ & $0.369$ & $0.895$ & $0.881$ \\
\makecell[l]{ASO}
& $\textbf{81.5\%}$ & $\textbf{0.430}$ & $\textbf{0.815}$ & $\textbf{0.813}$
& $\textbf{91.4\%}$ & $\textbf{0.287}$ & $\textbf{0.926}$ & $\textbf{0.916}$
& $\textbf{85.0\%}$ & $\textbf{0.357}$ & $0.824$ & $0.837$
& $\textbf{69.7\%}$ & $\textbf{0.520}$ & $\textbf{0.796}$ & $\textbf{0.737}$
& $\textbf{86.7\%}$ & $\textbf{0.387}$ & $\textbf{0.898}$ & $\textbf{0.885}$ \\
\bottomrule
\end{tabular}}
\caption{Main Results on \dataset. We divided all models into four groups and uniformly used Acc@0.5, MAE, PLCC, and SRCC to evaluate them across five quality dimensions.}
\label{tab:exp:results:main}
\end{table*}

\begin{table}[t]
\centering
\scalebox{0.7}{
\begin{tabular}{l|cccc|cc}
\toprule
\multirow{2}{*}{\textbf{Model}} & \multicolumn{2}{c}{\textbf{LSVQ-1080p}} & \multicolumn{2}{c|}{\textbf{KoNViD-1k}} & \textbf{VideoPhy2} & \textbf{MJ-Video} \\
 & \textbf{SRCC} & \textbf{PLCC} & \textbf{SRCC} & \textbf{PLCC} & \textbf{Acc (\%)} & \textbf{Acc (\%)} \\
\midrule
\multicolumn{7}{l}{\textit{ VLM-APIs}} \\
GPT-4.1 & 0.651 & 0.677 & 0.608 & 0.623 & 26.1 & 44.7 \\
Gemini-2.5Pro & 0.682 & 0.691 & 0.676 & 0.681 & 28.8 & 46.9 \\
\midrule
\multicolumn{7}{l}{\textit{ Open-source VLMs}} \\
Qwen2.5-VL & 0.624 & 0.610 & 0.658 & 0.642 & 20.2 & 34.6  \\
InternVL-3.5 & 0.610 & 0.595 & 0.635 & 0.620 & 18.1 & 38.5  \\
\midrule
\multicolumn{7}{l}{\textit{Specialized VQA Models}} \\
Q-Align & \textbf{0.797} & \textbf{0.830} & \textbf{0.865} & \textbf{0.877} & \textbf{23.2} & \textbf{22.0}\\
% VideoScoreV2 & 0.845 & 0.852 & 0.890 & 0.895 & \textbf{38.6} & \textbf{65.8}\\
VideoScoreV2 & - & - & - & - & \textbf{38.6} & \textbf{65.8}\\
\midrule
\multicolumn{7}{l}{\textit{Ours}} \\
Ours (SFT) & 0.765 & 0.737 & 0.781 & 0.813 & 30.6 & 50.8 \\
Ours (ASO) & 0.771 & 0.824 & 0.801 & 0.835 & 34.4 & 55.2\\
\bottomrule
\end{tabular}
}
\caption{Cross-benchmark evaluation. We report SRCC/PLCC for VQA datasets and Accuracy (\%) for VideoPhy2/MJ-Bench-Video. Symbol `-' denotes results not reported in the original literature.}
\label{tab:cross_benchmark_v2}
\end{table}
\subsection{Experimental Setting}

\textbf{Evaluation Metrics.}
We evaluate score prediction accuracy using both correlation-based and error-based criteria.
Let $y_i^{(d)} \in \{0,1,\dots,5\}$ denote the ground-truth score of video $i$ on dimension $d$, and let $\hat{y}_i^{(d)}$ be the model prediction.

\begin{itemize}
    \item \textbf{Accuracy (Acc).}
    Since labels are discrete and derived by rounding a mean subjective score.
    Therefore, rather than strict exact-match accuracy, we report a tolerance-based accuracy Acc@0.5:
    \[
    \text{Acc}^{(d)}=\frac{1}{N}\sum_{i=1}^{N}\mathbb{I}\big[\lvert\hat{y}_i^{(d)} - y_i^{(d)}\rvert \leq 0.5 \big].
    \]
    This criterion treats predictions within half a level as correct, which is equivalent to rounding $\hat{y}_i^{(d)}$ to the nearest discrete level before comparison, and better reflects the practical uncertainty of subjective ratings.

    \item \textbf{Spearman Rank Correlation (SRCC).}
    SRCC measures the monotonic association between predictions and ground-truth rankings and is invariant to any strictly monotonic re-scaling of scores.
    We compute SRCC following Spearman's original definition~\cite{spearman1961proof}.

    \item \textbf{Pearson Linear Correlation (PLCC).}
    PLCC captures linear agreement between predictions and ground truth~\cite{pearson1895vii}.

    \item \textbf{Mean Absolute Error (MAE).}
    MAE quantifies absolute deviation on the original rating scale:
    \[
    \text{MAE}^{(d)}=\frac{1}{N}\sum_{i=1}^{N}\big|\hat{y}_i^{(d)}-y_i^{(d)}\big|.
    \]
\end{itemize}

\textbf{Implementation Details.}
We use a unified training and inference setup for all open-source VLM baselines and our method.
For each video, we uniformly set FPS is 2.0 and use a shared instruction template that requests both multi-dimensional scores and a short rationale.
For inference, we parse the predicted scores from the formatted output.

\textbf{SFT.}
We fine-tune Qwen2.5-VL-7B and other open-source VLMs with AdamW using a learning rate $\eta_{\text{SFT}}=\text{1e-6}$, weight decay $\text{0.01}$ for 1 epoch.
We train with global batch size $B=\text{32}$ (via gradient accumulation), and bfloat16 training.

\textbf{GRPO and ASO.}
Starting from the SFT checkpoint, we perform RL post-training using Analytic Score Optimization (ASO).
We use a KL regularization term to constrain the policy from drifting from the SFT model, with coefficient $\beta=\text{0.1}$, and sample $\text{8}$ rollouts per prompt for policy updates.
Additional hyperparameters (reward scaling, clip range, etc.) are provided in Appendix~\ref{app:sec:exp}.

\textbf{Cross-benchmark evaluation.}  
We conduct cross-benchmark experiments on widely used dataset LSVQ~\cite{ying2021patch}, KoNViD-1k~\cite{hosu2017konstanz}, VideoPyh2-test~\cite{bansal2025videophy}, and MJ-Video~\cite{tong2025mj}.
We evaluate all models using the same instruction template and deterministic decoding whenever supported. 
The details of how to map the benchmark’s ground-truth MOS to the corresponding evaluation target are shown in Appendix~\ref{app:sec:exp}.

\subsection{Baselines}
To comprehensively evaluate the effectiveness of our method, we compare UltraVQA-ASO against three categories of state-of-the-art models: (1) Closed-source VLM APIs, specifically GPT-4.1~\cite{openai2025gpt41} and Gemini-2.5Pro~\cite{gemini2025gemini25}, which serve as the upper-bound reference for general multimodal reasoning; (2) Open-source Generalist VLMs, including Qwen2.5-VL~\cite{qwen25vl}, MiniCPMV-4.5~\cite{minicpmv45}, VideoLLaMA3~\cite{zhang2025videollama}, VideoChat-Flash~\cite{li2024videochat}, InternVL-3.5~\cite{wang2025internvl3},and InternVideo-2.5~\cite{wang2025internvideo2} representing strong open-weight foundation models without VQA-specific alignment; and (3) Specialized VQA Models, namely FineVQ~\cite{duan2025finevq}, Q-Align~\cite{wu2023qalign}, and VideoScoreV2~\cite{videoscore2}.

\subsection{Main Results}
Table~\ref{tab:exp:results:main} compares UltraVQA-ASO with state-of-the-art closed-source APIs, open-source VLMs, and specialized VQA models.

\textbf{Superior Calibration over Generalist VLMs.} 
While general-purpose models (e.g., GPT-4.1, Qwen2.5-VL) possess strong semantic understanding, they struggle to map visual perceptions to the specific ordinal scale of VQA. UltraVQA-ASO effectively bridges this gap, leveraging the base model's knowledge while strictly aligning its scoring behavior to human standards, thereby achieving SOTA performance across all five dimensions.

\textbf{Robustness and Generalization.} 
Against specialized VQA models like Q-Align, ASO not only matches their scoring accuracy but significantly excels in complex semantic dimensions (e.g., Content Quality). Crucially, this advantage extends beyond in-domain data. As shown in Table~\ref{tab:cross_benchmark_v2}, ASO demonstrates robust cross-benchmark generalization. It significantly outperforms generalist backbones on physical reasoning (VideoPhy2) and preference tasks (MJ-Video) while maintaining competitive alignment against specialized VQA models. This confirms that our analytic objective fosters robust representation learning rather than merely overfitting to the UltraVQA distribution.

\subsection{Ablation Study}
To validate the effectiveness of our pipeline, we perform a step-wise decomposition in Table~\ref{tab:exp:results:main}, comparing SFT, GRPO, and ASO.

\textbf{The Necessity of Preference Alignment.} 
The SFT baseline yields suboptimal ranking performance, indicating that standard teacher-forcing is insufficient for capturing the fine-grained ordinal nature of quality judgments. Introducing preference alignment (via GRPO or ASO) leads to a significant performance leap. This confirms that explicitly optimizing for the score distribution—rather than just mimicking token probabilities—is critical for calibrating the model to human ranking criteria.

\textbf{Analytic vs. Stochastic Optimization.} 
Comparing alignment strategies, ASO consistently outperforms GRPO, particularly on dynamic dimensions like Motion Quality. We attribute this to the analytic nature of ASO (Eq. 10). Unlike GRPO, which relies on high-variance stochastic sampling to estimate gradients, ASO directly optimizes towards a theoretically derived optimal distribution using the entire probability mass. This allows ASO to capture subtle ordinal distinctions more effectively than the sparse reward signals in standard RL, providing a more stable and efficient alignment alternative.
\section{Conclusion}
\label{sec:conclustion}
In this paper, we introduce \textbf{UltraVQA}, a large-scale multi-dimensional benchmark characterized by five complementary quality dimensions and fine-grained sub-attribute tags, rigorously supported by multi-rater annotations and distilled rationale supervision. Complementing this resource, we propose \textbf{Analytic Score Optimization (ASO)}, an RL-inspired objective specifically tailored to discrete, ordinal score prediction. By deriving a closed-form optimal score distribution under KL regularization and training via soft-target imitation, ASO effectively addresses the discrete nature of quality scoring while avoiding the instability of online RL. Empirically, our 7B VLM instantiated with ASO demonstrates consistent improvements over strong open-source baselines and competitive performance against specialized VQA models, further validated by robust results on cross-benchmark transfer and calibration analyses.


% \bibliography{example_paper}
\bibliography{main}
\bibliographystyle{icml2026}

%%%%%%%%%%%%%%%%%%%%%%%%%%%%%%%%%%%%%%%%%%%%%%%%%%%%%%%%%%%%%%%%%%%%%%%%%%%%%%%
%%%%%%%%%%%%%%%%%%%%%%%%%%%%%%%%%%%%%%%%%%%%%%%%%%%%%%%%%%%%%%%%%%%%%%%%%%%%%%%
% APPENDIX
%%%%%%%%%%%%%%%%%%%%%%%%%%%%%%%%%%%%%%%%%%%%%%%%%%%%%%%%%%%%%%%%%%%%%%%%%%%%%%%
%%%%%%%%%%%%%%%%%%%%%%%%%%%%%%%%%%%%%%%%%%%%%%%%%%%%%%%%%%%%%%%%%%%%%%%%%%%%%%%
\newpage
\appendix
\onecolumn
\section{Detailed Derivation of ASO}
\label{app:sec:method}

We start with the constrained optimization problem defined in Eq. 6 of the main text:
\begin{equation}
    \max_{\pi} \mathbb{E}_{x \sim \mathcal{D}, s \sim \pi(\cdot|x)} [R(s, s^*)] \quad \text{s.t.} \quad D_{KL}(\pi(\cdot|x) || \pi_{\text{ref}}(\cdot|x)) \leq \epsilon
\end{equation}
To solve this, we introduce a Lagrange multiplier $\lambda > 0$ and form the Lagrangian:
\begin{equation}
    \mathcal{L}(\pi, \lambda) = \mathbb{E}_{s \sim \pi} [R(s, s^*)] - \lambda \left( D_{KL}(\pi || \pi_{\text{ref}}) - \epsilon \right)
\end{equation}
Expanding the KL divergence term:
\begin{equation}
    \mathcal{L}(\pi, \lambda) = \sum_s \pi(s) R(s) - \lambda \sum_s \pi(s) \log \frac{\pi(s)}{\pi_{\text{ref}}(s)} + \lambda \epsilon
\end{equation}
Taking the functional derivative with respect to $\pi(s)$ and setting it to zero:
\begin{equation}
    \frac{\partial \mathcal{L}}{\partial \pi(s)} = R(s) - \lambda (\log \pi(s) + 1 - \log \pi_{\text{ref}}(s)) = 0
\end{equation}
Rearranging the terms, we obtain the optimal policy form:
\begin{equation}
    \log \pi^*(s) = \log \pi_{\text{ref}}(s) + \frac{R(s)}{\lambda} - 1
\end{equation}
Exponentiating both sides and introducing a normalization constant $Z(x)$ to ensure $\sum \pi^*(s) = 1$:
\begin{equation}
    \pi^*(s|x) = \frac{1}{Z(x)} \pi_{\text{ref}}(s|x) \exp\left(\frac{R(s, s^*)}{\lambda}\right)
\end{equation}
This concludes the derivation of the closed-form optimal solution used in Eq. 10.

\section{Dataset}
\label{app:sec:data}
\subsection{Human Annotation Protocol}
Annotator training and guideline iteration. 
All annotators completed a dedicated training session before entering the formal labeling stage. 
After training, we ran a pilot phase where annotators labeled 2K videos and provided brief score rationales. 
We aggregated these rationales to identify common perceptual cues and recurring ambiguities, and then updated the annotation manual accordingly. 
The refined guideline was used for the subsequent large-scale annotation.

Multi-rater annotation with rationales. 
In the final stage, each video was scored by at least three annotators, independently. 
For every rated video, each annotator also provided a short justification to support the assigned scores. This design both improves reliability through multi-rater aggregation and preserves human evidence that can later be used for interpretation and modeling.

To quantitatively validate the reliability of our subjective annotations, we conducted a rigorous Inter-Annotator Agreement (IAA) analysis.
As shown in Table~\ref{app:tab:data:iaa}, we measured consistency using both \textbf{Relaxed Match ($\mathcal{R}$)} (percentage of pairs with score difference $\le 1.0$) and \textbf{Krippendorff's Alpha ($\alpha$)}.
We observed a substantial improvement from the pilot phase to the final phase, attributed to our refined annotation guidelines and rigorous annotator training.
Ultimately, the final dataset achieves strong agreement ($\alpha > 0.75$) across all dimensions, confirming the robustness of our ground-truth labels despite the inherent subjectivity of the VQA task.

\subsection{GPT-4.1 Rationale Expansion}
Filtering noisy annotations. After collecting human scores and rationales, we performed a quality screening step to remove videos with clear annotation bias (e.g., unstable or inconsistent labeling patterns), so that only reliable human judgments are used for downstream rationale generation.

Rationale expansion conditioned on human evidence. For each retained sample, we provide the original video, the human-assigned scores, and the collected human rationales as inputs to GPT-4.1, and ask it to expand the rationales into more detailed, structured explanations. \textbf{We explicitly incorporate the fine-grained sub-attribute taxonomy and the aggregated human tags for all five quality dimensions into the prompt context.} Importantly, the model is instructed to remain consistent with the human scores and to ground the expanded rationale in observable cues from the video and the provided human notes.

\begin{figure*}[t]
\centering
\includegraphics[width=1.0\linewidth]{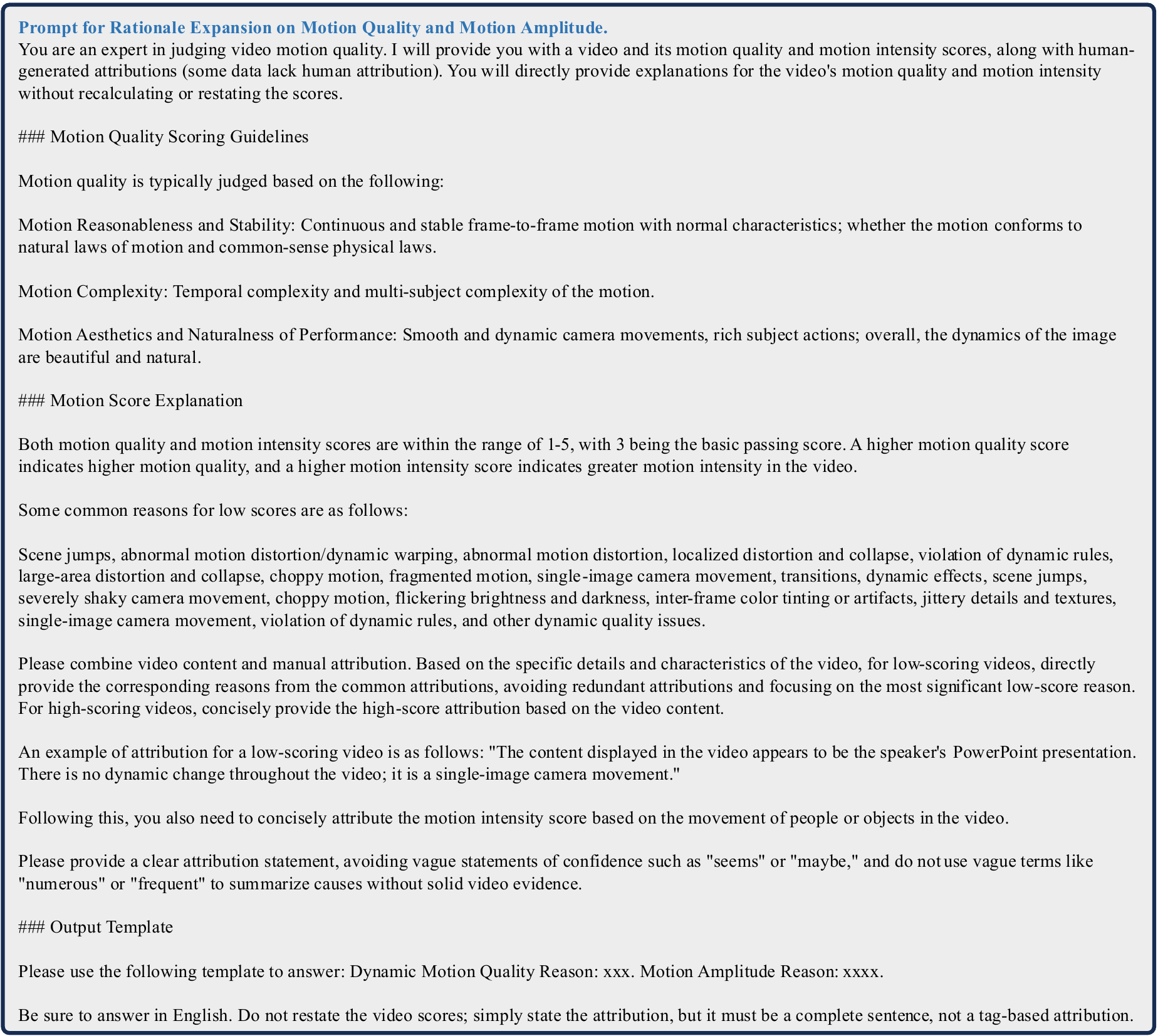}
\caption{
Prompt used for rationale expansion on Motion Quality and Motion Amplitude.
}
\label{fig:app:prompt:motion}
\end{figure*}

\begin{figure*}[t]
\centering
\includegraphics[width=1.0\linewidth]{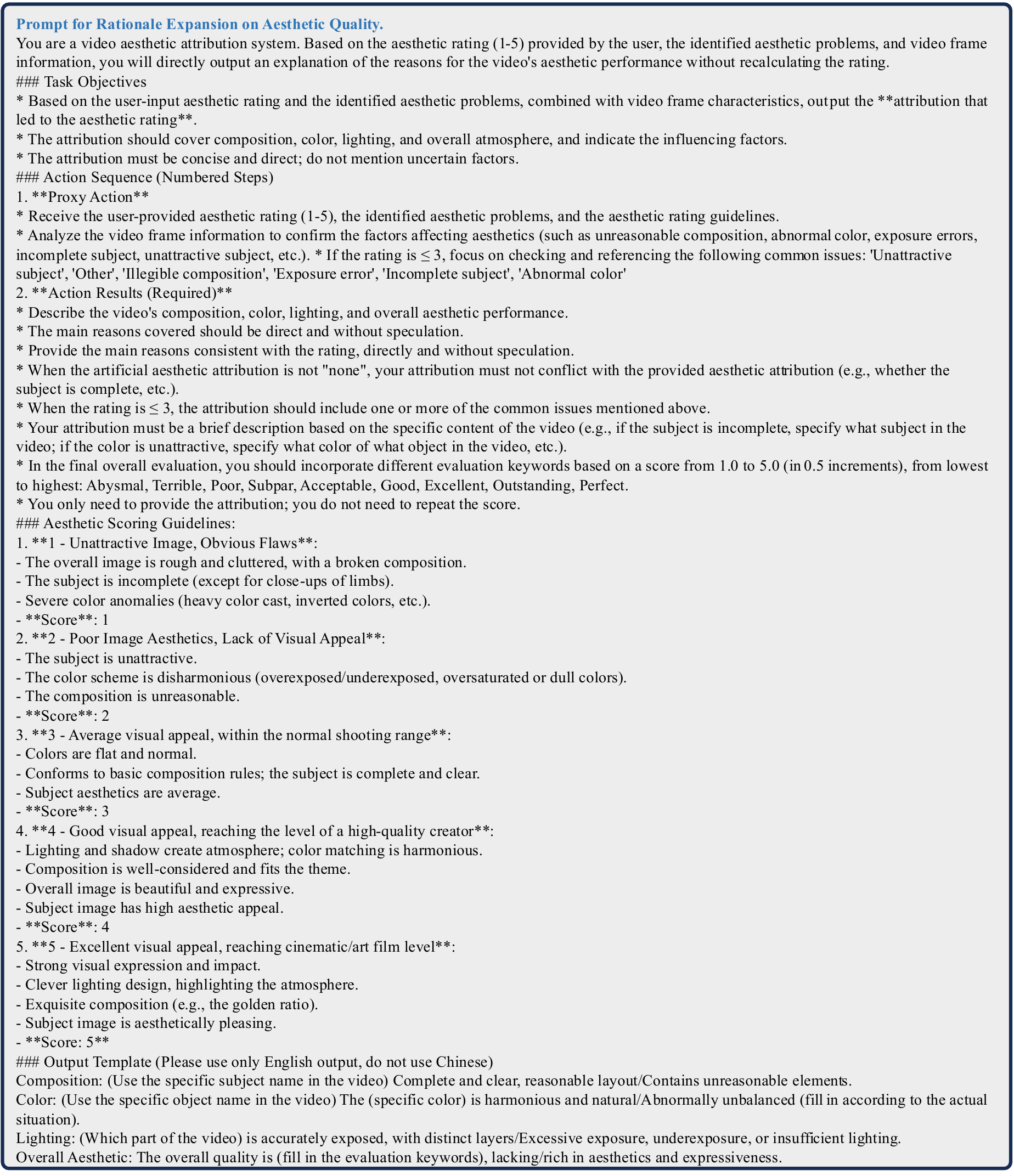}
\caption{
Prompt used for rationale expansion on Aesthetic Quality.
}
\label{fig:app:prompt:aes}
\end{figure*}

\begin{figure*}[t]
\centering
\includegraphics[width=0.85\linewidth]{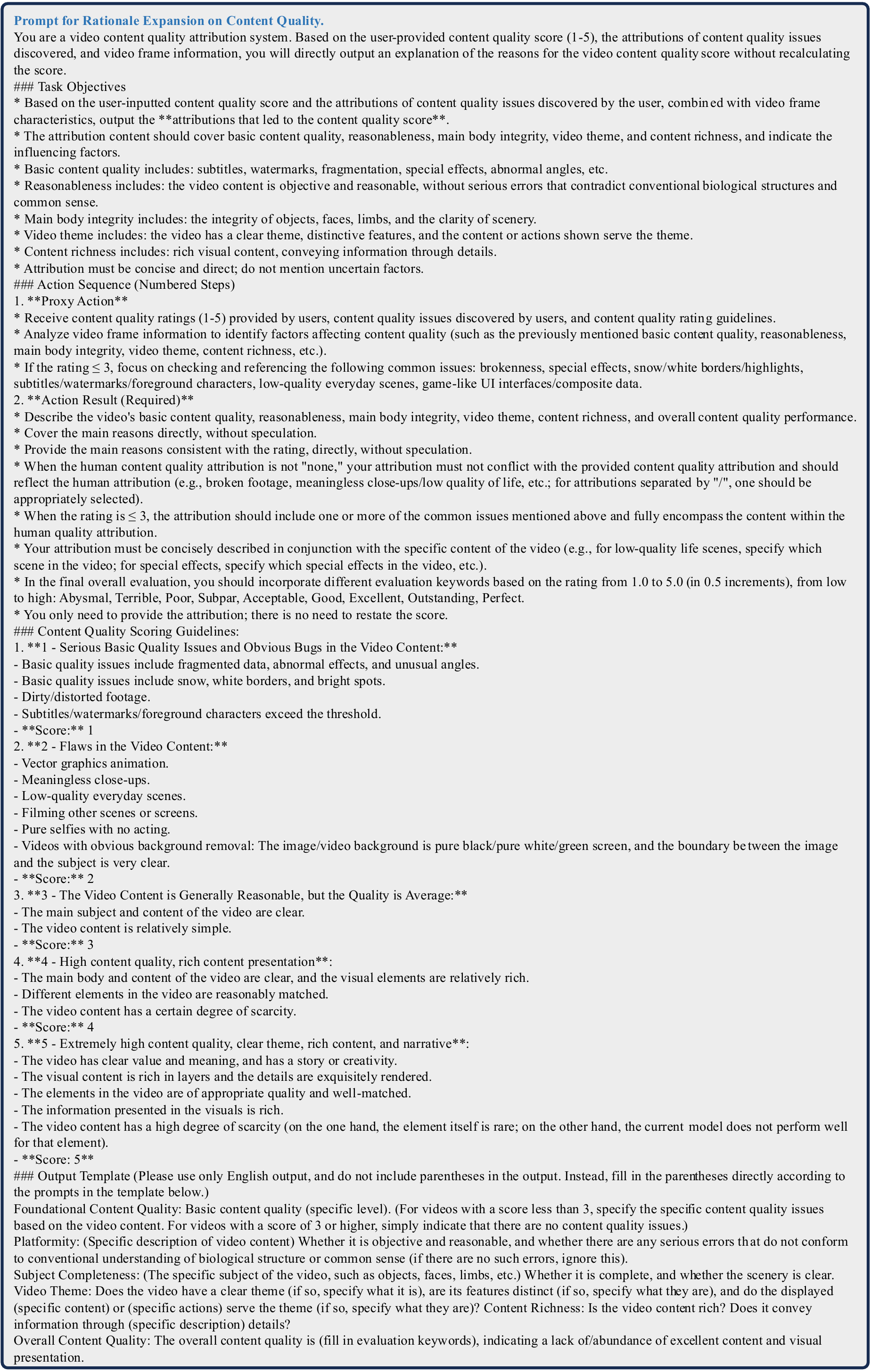}
\caption{
Prompt used for rationale expansion on Content Quality.
}
\label{fig:app:prompt:content}
\end{figure*}

\begin{figure*}[t]
\centering
\includegraphics[width=1.0\linewidth]{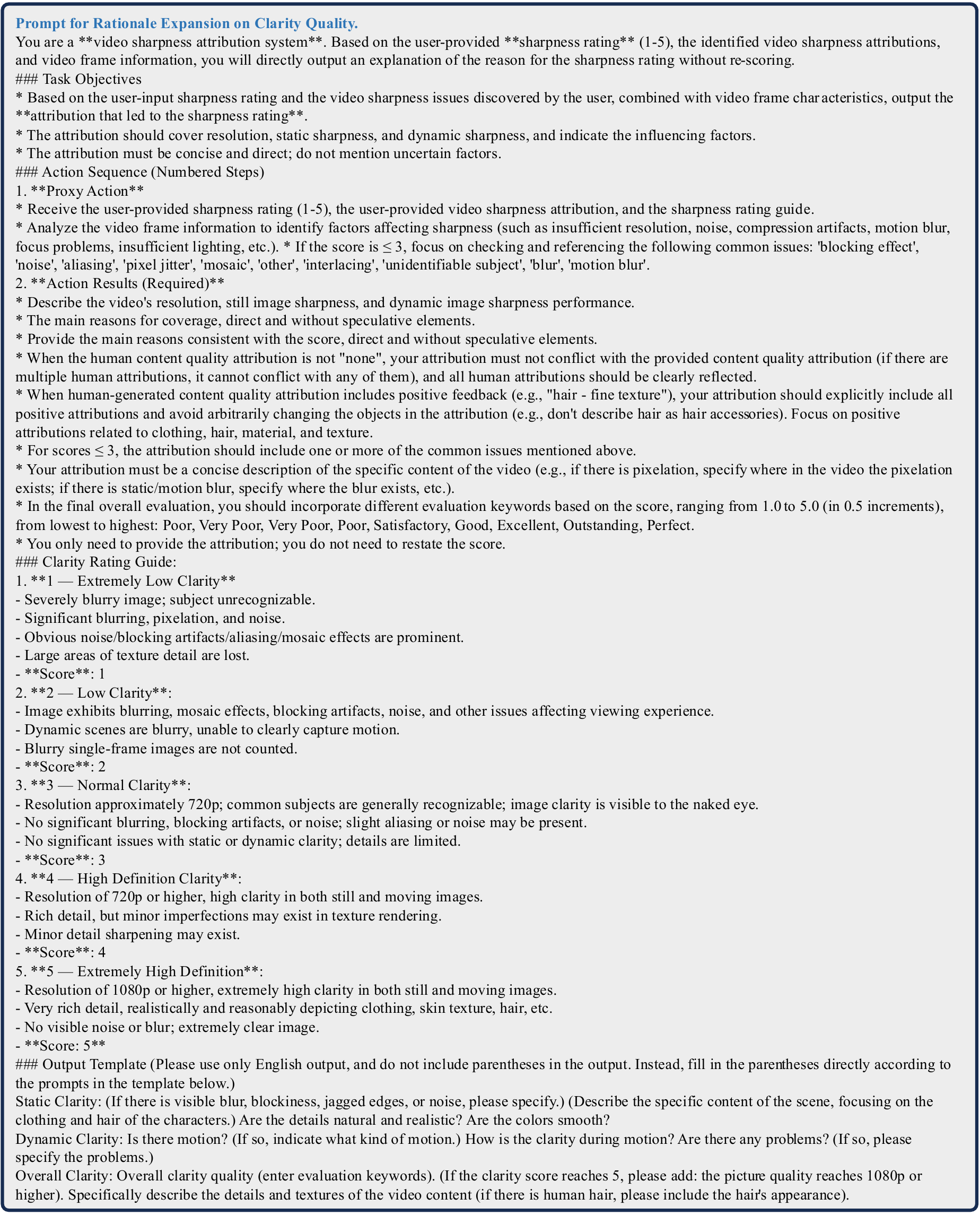}
\caption{
Prompt used for rationale expansion on Clarity Quality.
}
\label{fig:app:prompt:clarity}
\end{figure*}

Prompt selection via human review. We experimented with multiple prompt variants to control the format and style of the expanded rationales. Together with annotators, we reviewed the outputs and selected the final prompt used in all experiments, which is provided in Figure~\ref{fig:app:prompt:motion}, Figure~\ref{fig:app:prompt:aes}, Figure~\ref{fig:app:prompt:content}, and Figure~\ref{fig:app:prompt:clarity}.

\subsection{Rationale Vocabulary Analysis}
\begin{figure*}[t]
\centering
\includegraphics[width=1.0\linewidth]{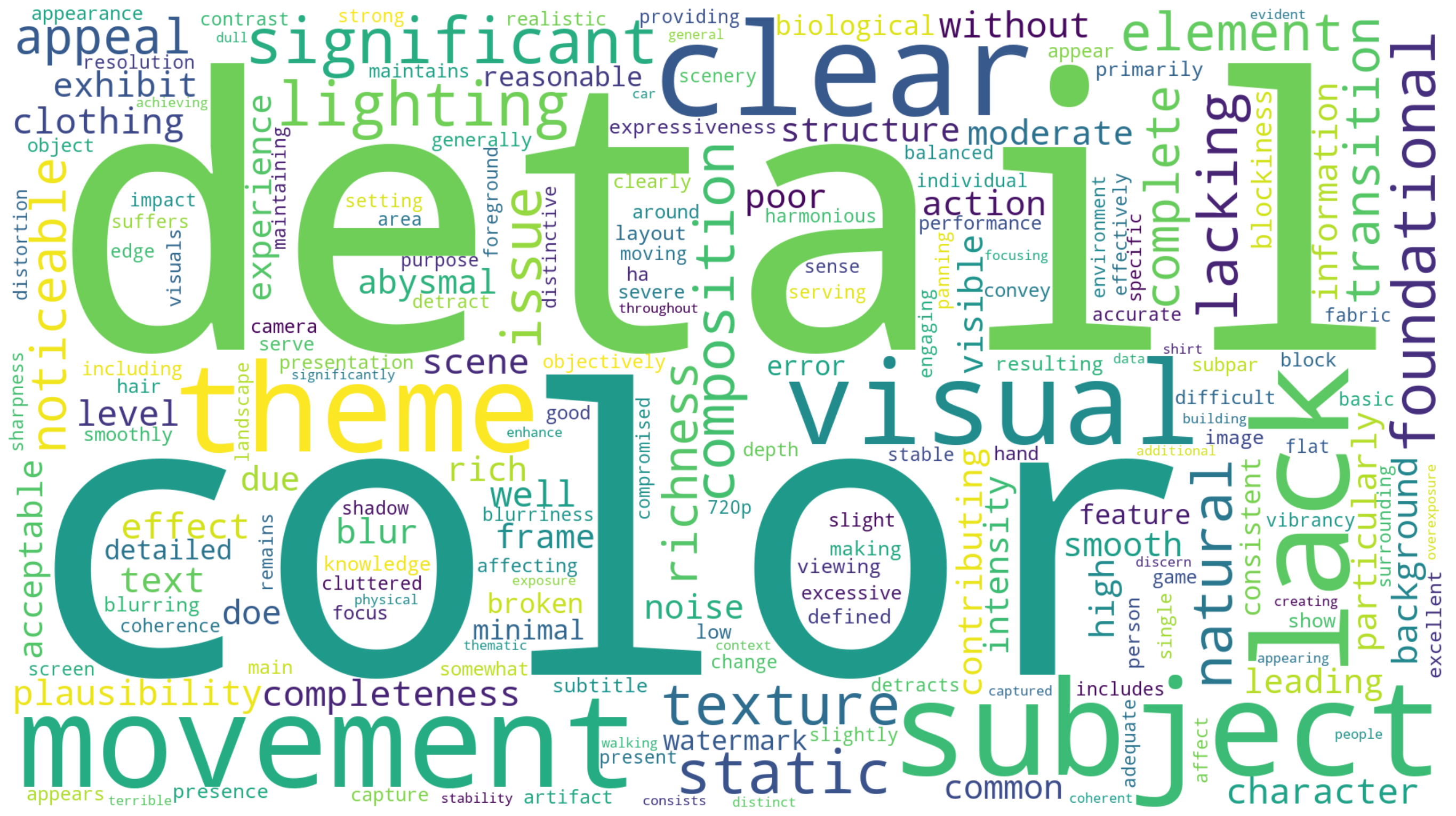}
\caption{
Word cloud display of rationale expansion.
}
\label{fig:app:data:wordcloud}
\end{figure*}

To visualize the semantic diversity of the synthesized rationales, we generated a word cloud from the expanded texts, as shown in Figure~\ref{fig:app:data:wordcloud}. Prior to visualization, we pre-processed the corpus by filtering out template-specific formatting tokens (e.g., "Reason:", "Output:") and common stop words. The resulting word cloud highlights that our method generates rich, diverse, and context-specific descriptive attributes across different quality dimensions.

\section{Experimental Details}
\label{app:sec:exp}
\subsection{Implementation Details}
\label{app:hyperparams}

\begin{table*}[t]
\centering
\resizebox{\textwidth}{!}{%
\begin{tabular}{l|ccccc}
\toprule
\multirow{2}{*}{\textbf{Annotation Phase}} & \textbf{Motion} & \textbf{Motion} & \textbf{Aesthetic} & \textbf{Content} & \textbf{Clarity} \\
 & \textbf{Quality} & \textbf{Amplitude} & \textbf{Quality} & \textbf{Quality} & \textbf{Quality} \\
 & ($\mathcal{R}$ / $\alpha$) & ($\mathcal{R}$ / $\alpha$) & ($\mathcal{R}$ / $\alpha$) & ($\mathcal{R}$ / $\alpha$) & ($\mathcal{R}$ / $\alpha$) \\
\midrule
Pilot Phase ($n = 2k$) & 85.2\% / 0.68 & 88.4\% / 0.71 & 81.5\% / 0.62 & 83.1\% / 0.65 & 89.0\% / 0.74 \\
\textbf{Final Phase} ($n = 40k$) & \textbf{92.4\% / 0.81} & \textbf{94.6\% / 0.85} & \textbf{88.7\% / 0.76} & \textbf{90.5\% / 0.79} & \textbf{96.2\% / 0.88} \\
\bottomrule
\end{tabular}%
}
\caption{Inter-Annotator Agreement (IAA) analysis across five quality dimensions. We report \textbf{Relaxed Match ($\mathcal{R}$)}, defined as the percentage of annotator pairs with score difference $\le 1.0$, and \textbf{Krippendorff's Alpha ($\alpha$)} to measure reliability. The significant improvement from the Pilot to the Final phase demonstrates the effectiveness of our refined guidelines and annotator training.}
\label{app:tab:data:iaa}
\end{table*}

\textbf{SFT Stage.} We fine-tune the Qwen2.5-VL-7B backbone using full parameter SFT. 
The learning rate is set to $1e-6$ with a cosine decay scheduler. 
The global batch size is 32. We train for 1 epoch to preserve the pre-trained knowledge.

\textbf{ASO Stage.} We initialize the model from the SFT checkpoint.
\begin{itemize}
    \item \textbf{KL Penalty ($\lambda$):} We set $\lambda = 1.0$. A larger $\lambda$ forces the model closer to the reference policy, while a smaller $\lambda$ allows more aggressive optimization towards the reward.
    \item \textbf{Reward Scale ($\beta$):} In the reward function $R(s, s^*) = -\beta |s - s^*|$, we set $\beta = 1.0$ (normalized such that the max reward difference is meaningful).
    \item \textbf{Optimization:} We use AdamW optimizer with learning rate $5e-6$ and batch size 32.
\end{itemize}

\textbf{Benchmark Mapping.} For LSVQ and KoNViD-1k, the ground-truth MOS (typically 0-5, 1-5 or 1-100) is normalized to our 1-5 scale for metric calculation. For VideoPhy2 and MJ-Bench, we use the accuracy metric definitions provided in their respective official repositories.
\section{Appendix E. Limitations}
\label{app:sec:limitations}

While UltraVQA-ASO demonstrates strong performance, we acknowledge two limitations:
(1) \textbf{Dependence on SFT Quality:} ASO operates by re-weighting the reference policy ($\pi_{ref}$). If the initial SFT model has zero probability mass on the correct score (e.g., due to severe overfitting), ASO cannot recover the correct prediction solely through re-weighting.
(2) \textbf{Discrete vs. Continuous:} Our derivation assumes a discrete label space (1-5). While this fits most human annotation protocols, extending ASO to continuous-valued regression tasks requires discretization, which may introduce quantization errors. Future work will explore continuous extensions of the analytic objective.
\end{document}